\pdfoutput=1

\documentclass[11pt]{article}

\usepackage[final]{acl/acl}

\usepackage{times}
\usepackage{latexsym}

\usepackage[T1]{fontenc}
\usepackage[utf8]{inputenc}

\usepackage{microtype}

\usepackage{inconsolata}

\usepackage{graphicx}

\PassOptionsToPackage{numbers, compress}{natbib}
\usepackage{natbib}

\usepackage{hyperref}
\usepackage{url}

\usepackage[utf8]{inputenc} % allow utf-8 input
\usepackage[T1]{fontenc}    % use 8-bit T1 fonts
\usepackage{hyperref}       % hyperlinks
\usepackage{url}            % simple URL typesetting
\usepackage{booktabs}       % professional-quality tables
\usepackage{amsfonts}       % blackboard math symbols
\usepackage{nicefrac}       % compact symbols for 1/2, etc.
\usepackage{microtype}      % microtypography
\usepackage{xcolor}         % colors
\usepackage{amsmath}
\usepackage{dsfont}
\usepackage{graphicx}
\usepackage{subcaption}
\usepackage{listings}
\usepackage{wrapfig}
\usepackage{lipsum}
\usepackage{arydshln}
\usepackage{enumitem}
\usepackage{xspace}
\usepackage{color, colortbl}
\usepackage{amssymb}
\usepackage{inconsolata}
\usepackage{multirow}
\usepackage{pifont}
\usepackage{moresize}
\usepackage[frozencache=true,cachedir=minted-cache]{minted} 

\newif\ifcomments
\commentstrue 
\ifcomments
    \providecommand{\jens}[2][]{{\protect\color{blue}{[Jens:\textbf{#1} #2]}}}
    \providecommand{\albert}[2][]{{\protect\color{magenta}{[Albert:\textbf{#1} #2]}}}
    \providecommand{\alex}[2][]{{\protect\color{orange}{[Alex:\textbf{#1} #2]}}}
    \providecommand{\khanh}[2][]{{\protect\color{cyan}{[Khanh:\textbf{#1} #2]}}}        \providecommand{\karthik}[2][]{{\protect\color{red}{[Karthik:\textbf{#1} #2]}}}
    \providecommand{\todo}[2][]{{\protect\color{red}{[TODO:\textbf{#1} #2]}}}
    
\else
    \providecommand{\jens}[2][]{}
    \providecommand{\albert}[2][]{}
    \providecommand{\alex}[2][]{}
    \providecommand{\khanh}[2][]{}
    
    \providecommand{\todo}[2][]{}
    \providecommand{\karthik}[2][]{}
\fi
\usepackage{amsmath,amsfonts,bm}

\def\eqref#1{equation~\ref{#1}}
\def\1{\bm{1}}

\DeclareMathAlphabet{\mathsfit}{\encodingdefault}{\sfdefault}{m}{sl}
\SetMathAlphabet{\mathsfit}{bold}{\encodingdefault}{\sfdefault}{bx}{n}

\def\gA{{\mathcal{A}}}

\def\gS{{\mathcal{S}}}

\definecolor{same}{HTML}{8b8b8b}
\definecolor{down}{HTML}{1d7b21}
\definecolor{up}{HTML}{9d0000}
\definecolor{OracleRow}{HTML}{f0f0f0}
\definecolor{action}{HTML}{0000fe}
\definecolor{state}{HTML}{800080}
\definecolor{reward}{HTML}{27a027}
\definecolor{done}{HTML}{b38600}
\definecolor{citecolor}{HTML}{670067}
\definecolor{mypink}{HTML}{ef09ed}

\hypersetup{
  colorlinks,
  citecolor=citecolor,
  linkcolor=citecolor,
  urlcolor=blue}

\renewcommand{\sectionautorefname}{\S\kern-0.2em}
\renewcommand{\subsectionautorefname}{\S\kern-0.2em}

\renewcommand{\vec}[1]{\boldsymbol{#1}}

\newcommand{\lwm}{LWM\xspace}
\newcommand{\messenger}{\textsc{Messenger}\xspace}

\newcommand{\none}{\texttt{Observational}\xspace}
\newcommand{\standard}{\texttt{Standard}\xspace}
\newcommand{\gpthard}{\texttt{GPTHard}\xspace}
\newcommand{\ours}{\texttt{EMMA-\lwm{}}\xspace}
\newcommand{\oracle}{\texttt{OracleParse}\xspace}

\newcolumntype{g}{>{\columncolor{OracleRow}}c}
\newcolumntype{f}{>{\columncolor{OracleRow}}l}

\title{Language-Guided World Models \raisebox{-0.2cm}{\includegraphics[width=1cm, trim={13cm 13cm 13cm 13cm}, clip]{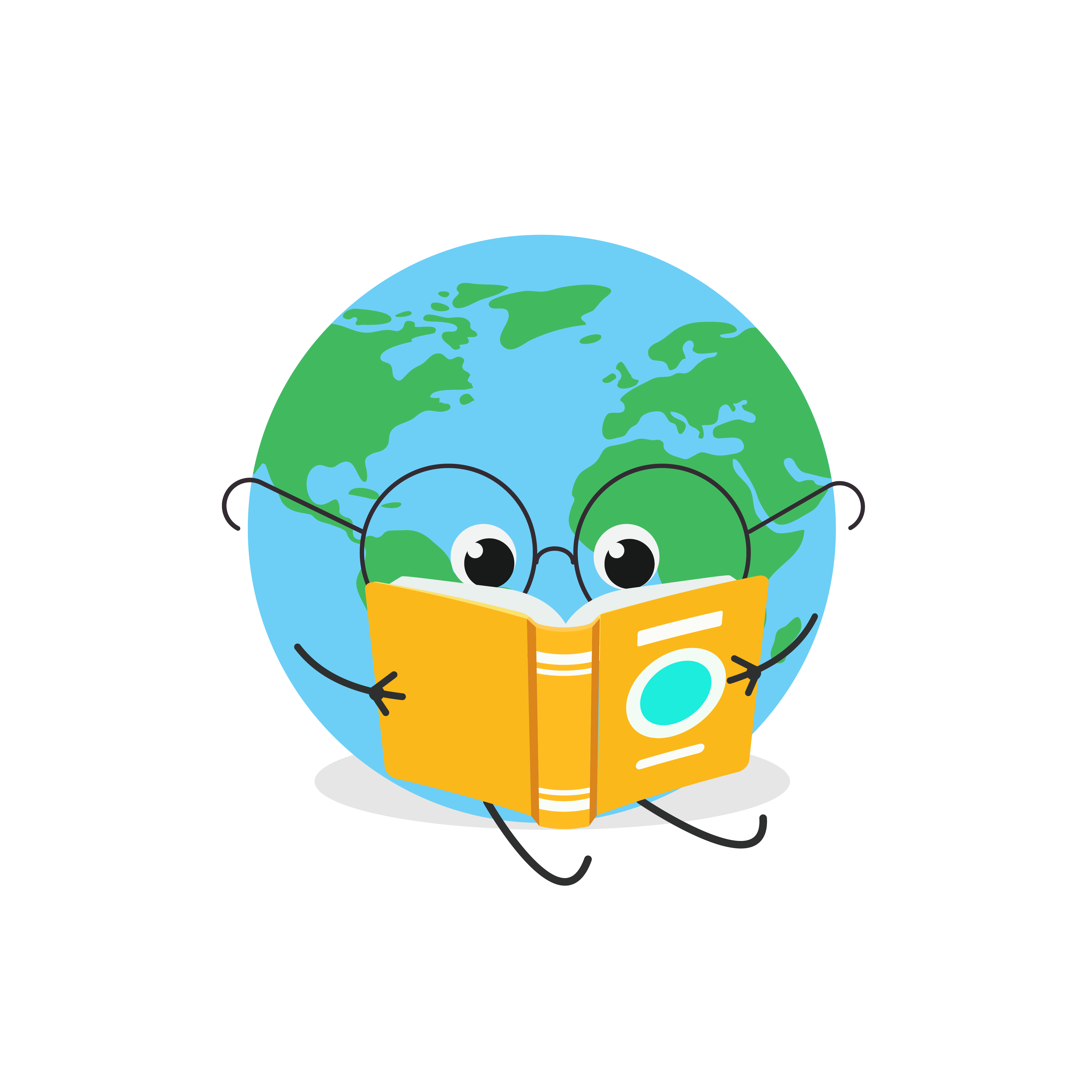}} \\ A Model-Based Approach to AI Control}

\author{\thanks{First two authors contribute equally. Correspondence email: \texttt{kxnguyen@berkeley.edu}.}Alex Zhang$^{\diamondsuit}$, \footnotemark[1]Khanh Nguyen$^{\spadesuit}$, Jens Tuyls$^{\diamondsuit}$, Albert Lin$^{\clubsuit}$,  Karthik Narasimhan$^{\diamondsuit}$ \\
$^{\diamondsuit}$ Princeton University \  
$^{\spadesuit}$University of California, Berkeley \\
$^{\clubsuit}$University of Southern California \\
\textbf{Project website}: \href{https://language-guided-world-model.github.io}{\textcolor{citecolor}{\texttt{language-guided-world-model.github.io}}}
}

\begin{document}

%\twocolumn[
\title{Language-Guided World Models \raisebox{-0.2cm}{\includegraphics[width=1cm, trim={13cm 13cm 13cm 13cm}, clip]{figures/globe_read_text.pdf}} \\ A Model-Based Approach to AI Control}

\maketitle
\begin{abstract}
This paper introduces the concept of \textit{Language-Guided World Models} (\lwm{}s)---probabilistic models that can simulate environments by reading texts. 
Agents equipped with these models provide humans with more extensive and efficient control, allowing them to simultaneously alter agent behaviors in \textit{multiple} tasks via natural verbal communication.
In this work, we take initial steps in developing robust LWMs that can generalize to compositionally novel language descriptions. 
We design a challenging world modeling benchmark based on the game of \messenger \citep{hanjie2021grounding}, featuring evaluation settings that require varying degrees of compositional generalization.
Our experiments reveal the lack of generalizability of the state-of-the-art Transformer model, as it offers marginal improvements in simulation quality over a no-text baseline. 
We devise a more robust model by fusing the Transformer with the EMMA attention mechanism \citep{hanjie2021grounding}. 
Our model substantially outperforms the Transformer and approaches the performance of a model with an oracle semantic parsing and grounding capability.
To demonstrate the practicality of this model in improving AI safety and transparency, we simulate a scenario in which the model enables an agent to present plans to a human before execution, and to revise plans based on their language feedback.
\looseness=-1
\end{abstract}

\begin{figure*}[t!]
    \centering
    \begin{subfigure}[b]{0.42\textwidth}
         \centering
         \includegraphics[width=\textwidth]{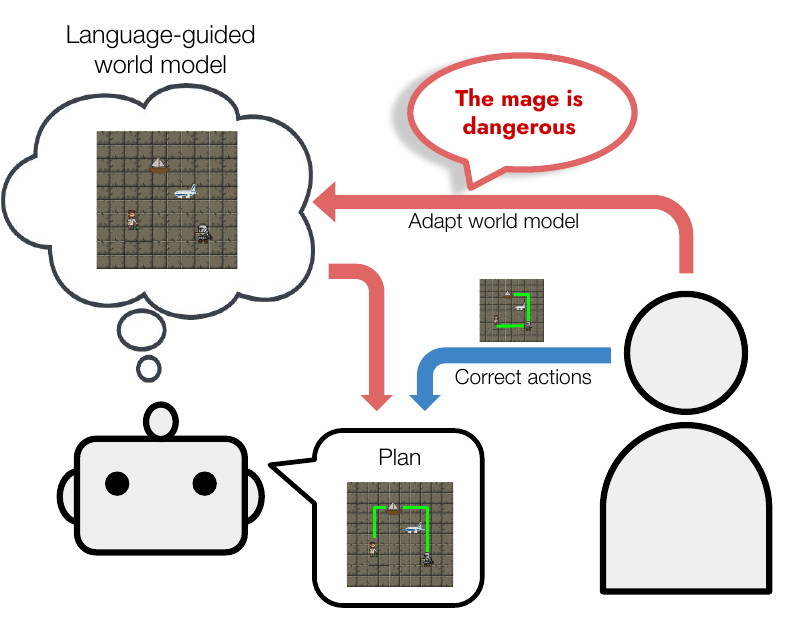}
         \caption{}
         \label{fig:application}
    \end{subfigure}
    % \hfill
    \begin{subfigure}[b]{0.5\textwidth}
         \centering
         \includegraphics[width=\textwidth]{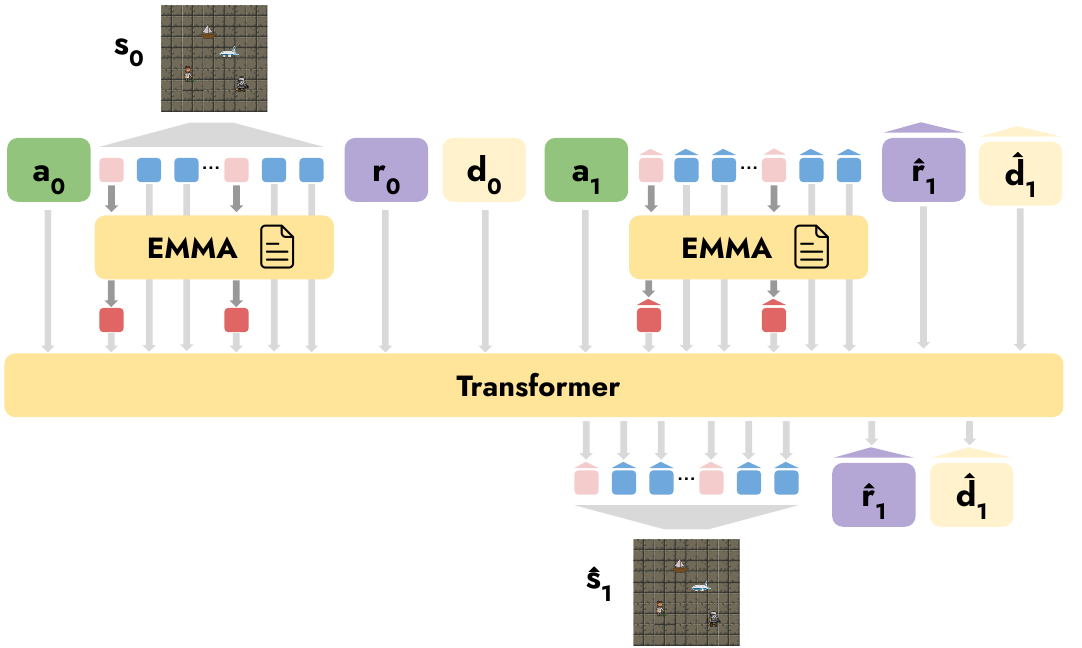}
         \caption{}
         \label{fig:model}
    \end{subfigure}
    \caption{Language-guided world models (LWMs) offer human an efficient mechanism to regulate artificial agents. (a) We illustrate a potential application of \lwm{}s to improving AI safety and transparency. These models enable an agent to generate visual plans and invite a human supervisor to validate them.
    Moreover, the human can adjust the plans by modifying the agent's world model with language feedback, in addition to directly correcting its policy. (b) We design an architecture for \lwm{}s that exhibits strong compositional generalization. We replace the cross-attention mechanism of the standard Transformer with a new attention mechanism inspired by \citet{hanjie2021grounding} to effectively incorporate language descriptions. We then train a model that auto-regressively generates tokenized observations conditioned on language descriptions and actions.  \looseness=-1 }
    \label{fig:teaser}
\end{figure*}

\section{Introduction}

\textit{Model-based agents} are artificial agents equipped with probabilistic ``world models'' that are capable of foreseeing the future state of an environment \citep{deisenroth2011pilco,schmidhuber2015learning}. 
World models endow these agents with the ability to plan and learn in imagination (i.e., internal simulation) and have led to exciting results in the field of reinforcement learning \citep{finn2017deep,ha2018world,chua2018deep,hafner2023mastering}.
These models have been studied extensively for the purpose of improving the autonomous performance of artificial agents. 

In this paper, we endorse and enhance the model-based approach for a different goal: to strengthen the controllability of artificial agents. 
Since all policies of a model-based agent are optimized with respect to a common world model, a human can adjust multiple policies simultaneously by making appropriate changes to this model.
This mechanism complements the model-free approach that updates policies individually, offering greater efficiency and flexibility in control.
For example, by incorporating the fact that the floor is slippery into the world model of a robot, a person can effectively remind it to handle \textit{every} object in a room with greater caution.
If the performance of the robot on a task remains unsatisfactory, the person can continue to fine-tune its policy for that specific task.
In contrast, without a world model, they have to separately adapt the robot's policies to the slippery-floor condition.
\looseness=-1

The model-based approach requires world models that can be easily modulated by humans.
Traditional world models fall short in this quality because they can only be modified using observational data, which is not a suitable medium for humans to convey intentions \citep{sumers2023show, zheng2023progressively}. 
To overcome the limitations of these models, we develop \textit{\textbf{L}anguage-Guided \textbf{W}orld \textbf{M}odels} (\lwm{}s)---world models that can be effectively steered through human verbal communication.
Agents equipped with \lwm{}s inherit all the benefits of model-based agents while being able to incorporate language-based supervision. 
This capability reduces human teaching effort and mitigates the risk of agents taking harmful actions in an environment to explore its dynamics. \lwm{}-based agents can also self-improve by reading ``free'' texts composed to guide humans (e.g., game manuals), reducing the subsequent effort to fine-tune them through direct interaction.

Building LWMs poses a unique research challenge: grounding language to environmental dynamics. 
This problem is difficult because the language used to describe environment dynamics can be incredibly rich and complex, encompassing a wide range of concepts such as entity names, appearances, motions, interactions, spatial and temporal relations, and more. Moreover, in natural settings, especially when describing artificial environments (e.g., games), new concepts are often introduced but may not always be clearly defined. 
Humans deal effectively with this issue because they possess remarkable reasoning capabilities that allow them to infer word meanings from observations. 
For example, a caption like ``\textit{the Ziff, which is chasing the player, is extremely hostile}'' and a video depicting this scene likely provide enough clues for a person to determine what ``the Ziff'' refers to, assuming that they are familiar with the concept of ``chasing''. 
Not only understanding word meanings, humans are also capable of applying newly learned words in novel ways, enabling imagination of new dynamics, such as envisioning a ``fleeing Ziff'' that runs away from the player.
\looseness=-1

Toward building world models with similar capabilities, we construct a benchmark based on the game of \messenger \citep{hanjie2021grounding}.
In this benchmark, a model is given trajectory ``videos'' of games involving several entities interacting with each other. 
Each video is accompanied by language descriptions of the attributes of the entities. 
The model begins with almost zero language understanding and has to identify the entities and learn the grounded meanings of their attributes purely by watching the videos.
At test time, it must demonstrate \textit{compositional generalization} by being able to simulate environments featuring entities with attributes different from those it observes during training.
For example, it has to portray a ``fleeing mage'' despite having only seen the mage chase the player in training games.
We design three evaluation settings that test for incrementally greater degree of compositional generalization.
\looseness=-1

Despite its apparent simplicity, our benchmark covers many complications in building robust \lwm{}s.
We find that the prominent Transformer model \citep{vaswani2017attention} struggles in the harder evaluation settings.
Even with a ground-truth disentangled representation of the observations, the model cannot learn generalizable grounding functions and yields minimal improvements in simulation quality compared to a model that ignores the language descriptions entirely.
We augment the model with the EMMA attention \citep{hanjie2021grounding}, which mimics a two-step reasoning process.
Our results confirm the effectiveness of this new architecture, as it robustly generalizes even in the hardest evaluation setting, outperforming baselines by substantial margins in various evaluation metrics. 
It is even competitive with a skyline model with an oracle semantic parsing and grounding capability. 
\looseness=-1

Last but not least, we illustrate a promising application of \lwm{}s by simulating a cautious agent that, instead of performing a task right away, uses its \lwm{} to generate an execution plan and asks a human to review it (\autoref{fig:application}).
This form of pre-execution communication can potentially improve the agent's safety and transparency, following the spirit of the guaranteed safe AI approach proposed by \citet{dalrymple2024towards}.
Moreover, it allows the human to improve the performance of the agent by revising the plan. 
In this setting, our \lwm-based agent has the advantage of being able to assimilate  \textit{language feedback} describing the environment dynamics. 
We demonstrate that the language understanding capabilities of our proposed \lwm are sufficient to enact this strategy. 
In the most challenging evaluation setting, without gathering additional interactions in the environment, the agent equipped with our model achieves an average reward three to four times higher than that of an agent using an observational world model.
\looseness=-1

We hope that our work will serve as a catalyst for exploring novel approaches to developing robust language-guided world models.
More generally, we call for the design of modular agents whose components are parameterized by natural language. 
As previously argued, a modular design can dramatically boost communication efficiency, because the same component may be involved in the learning of various policies.
We hypothesize that this approach can potentially surpass the efficiency of the currently prevalent approach that integrates language into a monolithic policy (e.g, \citet{bisk2016natural,misra2018mapping,anderson2018vision, narasimhan2018grounding,hanjie2021grounding,zhong2021silg} and work on large language models like \citet{ouyang2022training}).\looseness=-1

\section{Background: world models}

We consider a Markov Decision Process (MDP) environment $E$ with state space $\gS$, action space $\gA$, and transition function $M: \gS \times \gA \rightarrow \Delta(\gS \times \mathbb{R} \times \{0,  1\})$, where $\Delta$ denotes the set of all probability distributions over a set.
An agent implementing a policy $\pi(a \mid s): \gS \rightarrow \Delta(\gA)$ interacts with the environment by choosing actions using its policy.
Taking an action $a_{t} \sim \pi(s_t)$ in state $s_t$ transitions the agent to a new state $s_{t + 1}$, and incurs a reward $r_{t + 1}$ and a termination signal $d_{t + 1}$, where $s_{t + 1}, r_{t + 1}, d_{t + 1} \sim M(s_t, a_{t})$.

A (one-step) \textit{world model} $M_{\theta}$ \citep{robine2023transformerbased,micheli2022transformers,hafner2023mastering} is an approximation of $M(s_{t + 1}, r_{t + 1}, d_{t + 1} \mid s_t, a_{t})$.
A \textit{model-based agent} uses data gathered in the environment to construct a world model and leverages it to learn policies for accomplishing tasks.\footnote{Note that $M_{\theta}$ includes a reward function but can be combined with any other reward function for learning.} 
In contrast, a \textit{model-free} agent learns its policies directly from data collected in the environment. 

\paragraph{Model-based agents can require less effort to adapt.}
Because all policies of a model-based agent are derived from a shared world model, any modifications made to this model would affect all of them.
This feature can be exploited to reduce human effort in controlling this type of agent.
Specifically, suppose we concern $m$ tasks in the environment, necessitating $m$ policies.
If there is a change in the environment dynamics, a model-based agent only needs task-agnostic data to replicate this change in its world model. 
It can then re-optimize its policies with respect to the updated model. 
Meanwhile, a model-free agent needs to collect task-specific data to re-train all of its $m$ policies. 
The data collection cost of the model-free approach scales with $m$, whereas that of the model-based approach is independent of $m$, since the policy re-optimization step uses only data generated by the world model. 

\paragraph{Observational world models.} The dominant approach to world modeling learns a function $M_{\theta}(s_{t + 1}, r_{t + 1}, d_{t + 1} \mid h_t)$ parameterized by a neural network $\theta$ and conditioned on a history $h_t = (s_1, r_1, d_1, a_1, \ldots, s_t, r_t, d_t, a_t)$.
We refer to this class of models as \textit{observational world models} because they can be adapted with only observational data, through either in-weight learning (updating the model parameters to fit a dataset of observations), or in-context learning (plugging in a history of observations).

Relying on observation-based adaptation leads to two drawbacks. 
First, controlling these models is difficult because observations are inadequate for conveying complex, abstract human intentions.
Second, collecting observations requires taking real actions in the environment, which can be expensive, time-consuming, and risky. 

\section{Language-guided world models (\lwm{}s)}
\label{subsec:lwm}

We introduce \lwm{}s, a new class of world models that can interpret language descriptions to simulate environment dynamics.
These models address the drawbacks of observational world models. 
They allow humans to easily adapt their behavior through natural means of communication.
Consequently, humans can effectively assist these models, significantly reducing the amount of interactive experiences that they need to collect in environments.
In addition, these models can also leverage pre-existing texts written for humans, saving human effort to fine-tune them.  
\looseness=-1

\subsection{Formulation}

We consider a family of environments $E(\vec v)$ whose transition function has the form $M(s_{t + 1}, r_{t + 1}, d_{t + 1} \mid h_t, \vec v)$ where $\vec v$ is a parameter vector.
Plugging in a specific $\vec v$ gives rise to an environment.
We assume that each environment $E(\vec v)$ is accompanied by a \textit{language manual} $\vec \ell = ( l_1, \cdots, l_N )$ consisting of language descriptions $l_i$.
This manual describes $\vec v$ and the internal operations of $M$.
Our goal is to learn a world model $M_{\theta}(s_{t + 1}, r_{t + 1}, d_{t + 1} \mid h_t, \textcolor{purple}{\vec \ell})$ that approximates the true dynamics $M(s_{t + 1}, r_{t + 1}, d_{t + 1} \mid h_t, \vec v)$.
\looseness=-1

The training data for our \lwm{}s is a dataset $ \{(\tau^i, \vec \ell^i)\}$ where $\tau^i$ is a trajectory generated in an environment $E(\vec v_i)$ with $\vec v_i$ drawn from some distribution $P_{\textrm{train}}$, and $\vec \ell^i$ is the accompanying manual.  
Each trajectory $\tau = (s_1, r_1, d_1, a_1, \ldots, s_T, r_T, d_T)$ is a sequence of states, actions, rewards, and termination signals.
It can be viewed as a ``video'' that is annotated with actions and rewards. 
The trajectories are generated using a behavior policy, which can be a rule-based or learned policy, or a human.
\looseness=-1

\subsection{Modeling entity-based environments}

We view an environment as a set of $C$ \textit{entities} interacting with each other within a constrained space.
Each entity $c$ has a set of $K$ \textit{attributes}, each of which has value $v_k^c$.
There is a special attribute called the identity of the entity (e.g., the name of a character or object in a video game).
Each action triggers an event that changes a subset of attributes of a group of entities.
The specific change is determined by the attributes of the entities involved in the event (e.g., an enemy entity attacks a player when colliding with them).
In this work, we assume that each description in a manual portrays all attributes of an entity; hence, the number of descriptions $N$ is equal to $C$.

\paragraph{Testing for compositional generalization.}
With this formulation, the environment parameters $\vec v = (v_1^1, \cdots, v_K^1, v_1^2, \cdots, v^C_1, \cdots, v^C_K)$ is a vector that contains the attributes of the $C$ entities depicted in a manual. 
We are concerned with building \lwm{}s that, at test time, can simulate environments whose paramerer vectors are compositionally novel.
The term ``compositionally novel'' means that all components of the vector are individually seen during training, but the vector as a whole is previously unseen.
This implies that the manuals at test time are also new.

This problem requires a \lwm{} to be able to learn a representation of the transition function $M(\vec v)$ by studying the language of the manuals, and to extract the specific parameters $\vec v$ described by each manual.
The function $M(\vec v)$ has two important properties.
The first is the \textit{independence} among its parameters because they represent orthogonal attributes.
The second is the \textit{locality} of the parameters, as each is an attribute associated with only a single entity.
These properties make it difficult to recover the function exactly from purely observational data without injecting strong inductive biases into the learning model. 
\looseness=-1

\looseness=-1

\subsection{The \messenger-WM benchmark}

\begin{figure}[t!]
    \centering
    \includegraphics[width=\linewidth]{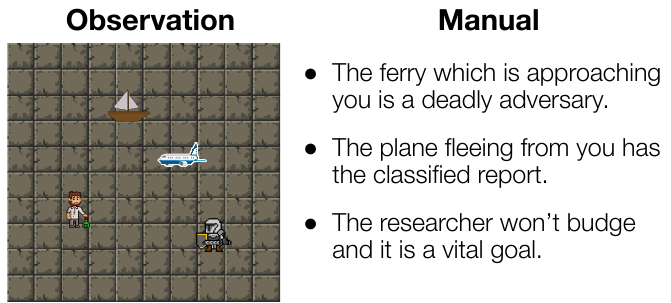}
    \caption{\messenger environment with  manual.\looseness=-1}
    \label{fig:messenger-example}
\end{figure}

The game of \messenger, developed by (\citet{hanjie2021grounding}; \autoref{fig:messenger-example}) exemplifies the class of environments discussed in the previous section. 
Despite being a simple grid-world environment, the dynamics possess the independence and locality properties that we want to study. 
In fact, it is our intention to use this visually simplistic environment to highlight the challenges in building \lwm{}s that are orthogonal to the computer graphics challenge of mapping state representations to realistic-looking outputs.
\looseness=-1

\paragraph{Environment dynamics.}
The game takes place in a $10 \times 10$ grid world.
A player interacts with entities of three \textit{roles}: message, goal, and enemy.
We use the stage-two version of the game, in which there are three entities, one of each role, in a game instance. 
In addition to the role, each entity is assigned an \textit{identity} among twelve possibilities (mage, airplane, orb, etc.) and a \textit{movement pattern} (chasing the agent, fleeing from the agent, immobile).
The objective of the player is to acquire the message and deliver it to the goal while avoiding the enemy. 
Fetching the message is awarded 0.5 points and delivering it to the goal adds another point.
If the player collides with the enemy or reaches the goal without carrying the message, the game ends, and the player receives -1 points.

\paragraph{Game manual.} 
A game's manual consists of three descriptions corresponding to the three entities.  
\messenger provides a dataset of 5,316 language descriptions, each of which describes a combination of identity, role, and movement. 
The descriptions employ various linguistic expressions for each identity, role, or movement pattern (e.g., an airplane can be mentioned as a ``plane'', ``jet'', or ``airliner''), making it non-trivial to interpret.

\paragraph{Evaluation settings.} 
To test for compositional generalization, we construct three evaluation settings, ordered in increasing degree of difficulty:
\begin{itemize}[leftmargin=*,nolistsep]
    \item \textbf{NewCombo (easy).} Each game features a combination of three identities that were never seen together in a training game. 
    However, the role and movement pattern of each identity are the same as during training.
    
    \item \textbf{NewAttr (medium).} The three identities were seen together in a training game, but each identity is assigned at least a new attribute (role, or movement pattern, or both). 
    
    \item \textbf{NewAll (hard).} This setting combines the difficulties of the previous two. The identity triplet is novel, and each identity is assigned at least a new attribute. 

\end{itemize}

To generate trajectories, we implement rule-based behavior policies that execute various intentions: act randomly, avoid the enemy, suicide (go to the enemy), obtain the message, and win the game (obtain the message and deliver it to the goal).
We generate a total of 100K trajectories for training, each of which is generated by rolling out a uniformly randomly chosen rule-based policy. 
More details of the data are given in \autoref{sec:dataset}. 
Our evaluation is more comprehensive than the original \messenger paper's evaluation, which does not construct different levels of compositional generalization, and is more difficult than the setting of \citet{lin2024learning}, which does not concern generalization.

To succeed in \messenger-WM, a model must be able to understand the non-trivial concepts represented by the attributes.
For example, the concept of ``chasing'' involves planning actions to reduce the distance between two entities. 
The model must also capture the independence of the attributes, despite observing correlations in the training data (e.g., the ``mage'' is never immobile during training).
Finally, to reflect the locality of the attributes, the model needs to learn a representation that disentangles the entities and to route attributes to the right entities. 
For example, the movement of one entity should not influence that of another.
These are among the difficult, under-explored problems in machine learning, making \messenger-WM a respectable research challenge.
We will empirically show that the state-of-the-art Transformer architecture struggles to perform well on the benchmark, suggesting that it may be insufficient for tackling more complex world-modeling problems.

\section{Modeling approach}
\label{sec:model}

\paragraph{State representation.}
In \messenger, a state $s$ is represented by an $H \times W$ grid with $C$ channels (an $H \times W \times C$ tensor), where each channel corresponds to an entity.  
In each channel $c$, there is a single non-zero cell $s(h, w, c)$ that represents the identity of the entity.
The position of this cell is the location of the entity in the grid. 
We note that this is an idealized representation that disentangles the entities.
Even so, the problem remains challenging, as the model needs to recognize attributes mentioned in the manual and associate them with the right entity token. 
This requires a special attention mechanism, which we will introduce shortly.
Meanwhile, learning entity-disentangled representations for pixel-based environments remain an open problem, which we defer to future work. 
\looseness=-1

\paragraph{World modeling as sequence generation.} 
Our model (illustrated in \autoref{fig:model}) is an encoder-decoder Transformer \citep{vaswani2017attention} which encodes a manual $\vec \ell$ and decodes a trajectory $\tau$. 
We transform the trajectory into a long sequence of tokens and train the model as a sequence generator. 

Concretely, our model processes a data point $(\tau, \vec \ell)$ as follows. 
For the manual $\vec \ell = \{ l_i \}_{i = 1}^N$, we first use a pre-trained BERT model to convert each description $l_i$ into a sequence of hidden vectors.
We feed each sequence to a Transformer encoder, which outputs a tensor $\vec m^{\textrm{enc}}$ of size $N \times L \times D$, where $N = C$ is the number of descriptions, $L$ is the maximum number of words in a description, and $D$ is the hidden size.

For the trajectory, we convert each tuple $(a_{t - 1}, s_t, r_t, d_t)$ into a token block $B_t$.
The first action $a_0$ is set to be a special \texttt{<s>} token.
Each state $s_t$ is mapped to $3C$ tokens $(i_t^1, h^1_t, w^1_t, \cdots, i^{C}_t, h^{C}_t, w^{C}_t)$, which represents each of the $C$ entities by its identity $i$ followed by its location $(h, w)$. 
The real-valued reward $r_t$ is discretized into an integer label, and the termination signal $d_t$ is translated into a binary label.
In the end, $B_t$ consists of $3C + 3$ tokens $(a_{t - 1}, i_t^1, h^1_t, w^1_t, \cdots, i^{C}_t, h^{C}_t, w^{C}_t, r_t, d_t)$.
Finally, we concatenate all $T$ blocks in the trajectory into a sequence of $T \times (3C + 3)$ tokens, embed them into a $T \times (3C + 3) \times D$ tensor, and add positional embeddings.
We will use bold notation (e.g., $\vec a$, $\vec i$) to refer to the resultant embeddings of the tokens. 

\paragraph{Entity mapper with multi-modal attention.} 
We implement a variant of EMMA (\citet{hanjie2021grounding}) that first identifies the description that mentions each entity and extracts from it words corresponding to the attributes of the entity.  
From the tensor $\vec m_n^{\textrm{enc}}$ computed by the encoder, we generate a key tensor $\vec m^{\textrm{key}}$ and a value tensor $\vec m^{\textrm{val}}$, both of which are of size $N \times L \times D$, where
\begin{align}
    \vec m^{\textrm{key}}_n &= \texttt{Softmax}( \texttt{Linear}_{\textrm{key}}( \vec m_{n}^{\textrm{enc}})^{\top})  \vec m_{n}^{\textrm{enc}} \nonumber \\
    \vec m^{\textrm{val}}_n &= \texttt{Softmax}( \texttt{Linear}_{\textrm{val}}( \vec m_{n}^{\textrm{enc}})^{\top})  \vec m_{n}^{\textrm{enc}} 
\label{eqn:attn-val}
\end{align} for $1 \leq n \leq N$. Here, $\texttt{Linear}^{D \rightarrow 1}_{\textrm{key}}$ and $\texttt{Linear}^{D \rightarrow 1}_{\textrm{val}}$ are linear layers that transform the input's last dimension from $D$ to 1, and $\texttt{Softmax}(\cdot)$ applies the softmax function to the last dimension. 
Intuitively, we want each $\vec m^{\textrm{key}}_{n}$ to retain words that signal the identity of the entity mentioned in the $n$-th description (e.g., \textit{ferry}, \textit{plane}, \textit{researcher}), and $\vec m^{\textrm{val}}_{n}$ to retrieve words depicting the other attributes (e.g., \textit{approaching}, \textit{deadly}, \textit{fleeing}).

Let $\vec i^c_t$ be the embedding of the identity of entity $c$. 
We perform a dot-product attention with $\vec i^c_t$ as the query, $\vec m^{\textrm{key}}$ as the set of keys, and $\vec m^{\textrm{val}}$ as the set of values to compute the attribute features of $c$
\begin{align}
    \vec z^c_t = \texttt{DotAttend}(\vec i^c_t, \vec m^{\textrm{key}}, \vec m^{\textrm{val}})
\end{align} The features are added to the identity tokens $\vec i^c_t$. The final input of the model is as follows:
\begin{align}
    (\vec a_{t - 1}, ( \textcolor{red}{\vec i_t^c + \vec z^c_t}, \vec h^c_t, \vec w^c_t )_{c = 1}^C, \vec r_t, \vec d_t)
\label{eqn:decoder-input}
\end{align}
Unlike the standard encoder-decoder Transformer, our architecture does not perform cross-attention between the encoder and the decoder because information from the encoder has already been incorporated into the decoder through EMMA.

\paragraph{Model training.} We train the model to minimize cross-entropy loss with respect to the ground-truth (tokenized) trajectories in the training set. 
The label at each output position is the next token in the ground-truth sequence. 
In particular, we do not compute the losses at the positions of the action tokens and the first block's tokens, because those tokens will be set during inference.
\section{Experiments}

\subsection{Baselines}

We compare our model, which we call \ours, with the followings:
\begin{enumerate}[label=(\alph*),nolistsep]
    \item \textbf{\none} world model does not leverage textual information. It is identical to \ours except that we zero out the manual representation $\vec m^{\textrm{enc}}$; 
    \item \textbf{\standard} is the encoder-decoder Transformer model following \citet{vaswani2017attention} with multi-headed cross-attention between the decoder and the encoder. Similarly to \ours, the model uses BERT to initially encode the manual into hidden vectors. The encoder applies self-attention to the hidden vectors of each description separately, instead of joining all vectors into a sequence and applying self-attention to it;\looseness=-1
    \item \textbf{\gpthard} is similar to \ours but uses ChatGPT instead of EMMA to ground descriptions to entities. More details about this model are in \autoref{appendix:chatgpt};
    \item \textbf{\oracle} is the same as \gpthard, but uses an oracle information extraction function. A description like \textit{``the crucial target is held by the wizard and the wizard is fleeing
from you''} is converted into \textit{``mage fleeing goal''} for this model. 
\looseness=-1

\end{enumerate} We train all models using AdamW \citep{loshchilov2017decoupled} for $10^5$ iterations. For further details, please refer to~\autoref{sec:training}.

\begin{table}[t]
    \setlength{\tabcolsep}{5pt}
    \small
    \caption{Cross entropy losses ($\downarrow$) of different models on test ground-truth trajectories. Note that the minimum loss is non-zero because the \messenger environment is stochastic. We run each model with five different random seeds, selecting the final checkpoint for each seed based on the loss in the development NewAll split. We report the mean losses with 95\% t-value confidence intervals. The bold number in each column indicates the best non-oracle mean. 
    }
    \label{tab:real-eval}
    \centering
    \begin{tabular}{lccc}
    \toprule
    & NewCombo & NewAttr & NewAll \\
    World model & (easy) & (medium) & (hard) \\
    \midrule
    \none & 0.12 \textcolor{gray}{\ssmall{$\pm$ 0.04}} & 0.18 \textcolor{gray}{\ssmall{$\pm$ 0.02}} & 0.19 \textcolor{gray}{\ssmall{$\pm$ 0.01}} \\
    \standard & 0.10 \textcolor{gray}{\ssmall{$\pm$ 0.04}} & 0.15 \textcolor{gray}{\ssmall{$\pm$ 0.04}} & 0.16 \textcolor{gray}{\ssmall{$\pm$ 0.03}} \\
    \gpthard & 0.10 \textcolor{gray}{\ssmall{$\pm$ 0.02}} & 0.15 \textcolor{gray}{\ssmall{$\pm$ 0.01}} & 0.16 \textcolor{gray}{\ssmall{$\pm$ 0.00}} \\
    \ours & \textbf{0.08 \textcolor{gray}{\ssmall{$\pm$ 0.01}}} & \textbf{0.10 \textcolor{gray}{\ssmall{$\pm$ 0.02}}} & \textbf{0.13 \textcolor{gray}{\ssmall{$\pm$ 0.01}}} \\
    \rowcolor{OracleRow}
    \oracle & 0.08 \textcolor{gray}{\ssmall{$\pm$ 0.01}} & 0.09 \textcolor{gray}{\ssmall{$\pm$ 0.02}} & 0.12 \textcolor{gray}{\ssmall{$\pm$ 0.06}} \\

    \bottomrule
    \end{tabular}    
\end{table}

\begin{figure*}[ht!]
    \centering
    \includegraphics[width=0.8\linewidth]{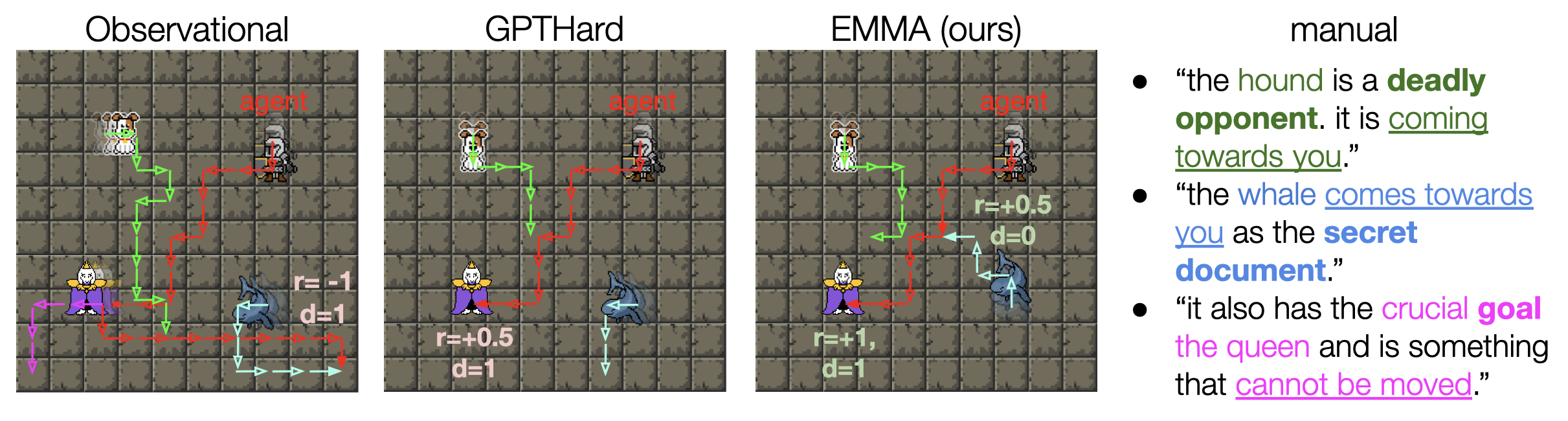}
    \caption{A qualitative example taken from the NewAll split. The \none model mistakenly captures the movement patterns of the \textcolor{mypink}{immobile queen goal} and the \textcolor{cyan}{chasing whale message}. It also misrecognizes the whale as an enemy, predicting a wrong reward $r$ and incorrectly predicting a termination state $d$ after the player collides with this entity.  The \gpthard model incorrectly identifies the queen as the message and predicts the whale to be fleeing. Meanwhile, our model \ours accurately captures all of those roles and movements.}
    \label{fig:qual-example}
    \vspace{-5pt}
\end{figure*}

\subsection{Results}

\paragraph{Evaluation with ground-truth trajectories.} \autoref{tab:real-eval} shows the cross-entropy losses of all models on ground-truth trajectories sampled from the true environment dynamics (more in \autoref{sec:extended-result}).
In the more difficult NewAttr and NewAll splits, our \ours model consistently outperforms all baselines, nearing the performance of the \oracle model. 
As expected, the \none model is easily fooled by spurious correlations between identity and attributes, and among attributes. 
A specific example is illustrated in \autoref{fig:qual-example}. 
There, the \none model incorrectly captures the movement of the whale and the queen. 
It also mistakenly portrays the whale as an enemy, whereas, in fact, the entity holds the message. 
In contrast, \ours is capable of interpreting the previously unseen manual and accurately simulates the dynamics.
\looseness=-1

The performance of the \standard model is sensitive to initialization; in some runs, it performs as well as \ours, but in others it performs as badly as \none.  
A plausible explanation is that the model's attention mechanism lacks sufficiently strong inductive biases to consistently find generalizable solutions. Our results agree with previous work on the lack of compositional generalizability of Transformers, which is often remedied by adding various forms of inductive bias \citep{keysers2019measuring,jiang2021inducing,chaabouni2021can,dziri2023faith}.

Another interesting finding is that the \gpthard model does not perform as well as expected.
As a reminder, this model relies on ChatGPT to parse identities from descriptions and only needs to learn to extract attributes.
Its underperformance compared to \ours can be attributed to (i) the imperfection of ChatGPT in identifying identities in descriptions (its accuracy is around 90\%; see \autoref{sec:dataset}) and (ii) the fact that \ours jointly learns to extract both identity and attribute words, which may be more effective than learning to extract only attribute words.

\begin{table*}[!htb]
    \setlength{\tabcolsep}{5pt}
    \scriptsize
    \caption{Results on imaginary trajectory generation. $\Delta_{\textrm{dist}}$  measures the similarity between the distances from the player to an entity in a real trajectory and the corresponding imaginary trajectory. The bold number in each column represents the best non-oracle result. \ours outperforms all baselines in all metrics.}
    \label{tab:imagine}
    \centering
    \begin{tabular}{lccccccccc}
    \toprule
    & \multicolumn{3}{c}{$\Delta_{\textrm{dist}} (\downarrow)$} & \multicolumn{3}{c}{Non-zero reward precision ($\uparrow$)} & \multicolumn{3}{c}{Termination precision ($\uparrow$)} \\
    & NewCombo & NewAttr & NewAll  & NewCombo & NewAttr & NewAll & NewCombo & NewAttr & NewAll\\
    World model & (easy) & (medium) & (hard)  & (easy) & (medium) & (hard) & (easy) & (medium) & (hard)\\
    \cmidrule(lr){1-1} \cmidrule(lr){2-4} \cmidrule(lr){5-7} \cmidrule(lr){8-10}
    \none & 2.04& 2.91& 3.00& 0.39& 0.20& 0.15& 0.51& 0.33& 0.28\\
    \standard & 0.82& 1.48& 1.68& 0.68& 0.43& 0.50& 0.75& 0.55& 0.62\\
    \gpthard & 0.89& 2.74& 2.89& 0.75& 0.34& 0.25& 0.79& 0.45& 0.45\\
    \ours & \textbf{0.57 }& \textbf{1.14}& \textbf{1.29}& \textbf{0.88}& \textbf{0.69}& \textbf{0.70}& \textbf{0.88}& \textbf{0.75}& \textbf{0.71}\\
    \rowcolor{OracleRow}\oracle & 0.49& 0.77& 0.92& 0.93& 0.81& 0.77& 0.89& 0.84& 0.79\\
    \bottomrule
    \end{tabular}    
\end{table*}

\paragraph{Evaluation with imaginary trajectories.} In this evaluation, for each world model and test trajectory, we reset the model to the initial state of the trajectory and sequentially feed the actions in the trajectory to the model until it predicts the end of the episode.
This process generates an imaginary trajectory.
We refer to the evaluation trajectory as the real trajectory.
We compute precisions of predicting non-zero rewards ($r \neq 0$) and terminations ($d = 1$). 
To evaluate movement prediction, we compare the distances from the player to an entity in the real and imaginary trajectories.
Concretely, let $\delta_{c, t}^{\text{real}}$ and $\delta_{c, t}^{\text{imag}}$ be the Hamming distances from the player to entity $c$ at the $t$-th time step in a real trajectory $\tau_{\textrm{real}}$ and an imaginary trajectory $\tau_{\textrm{imag}}$, respectively. 
We calculate the average difference in a specific time step: $\Delta_{\text{dist}} = \frac{1}{|\mathcal{D}_{\textrm{eval}}|} \sum_{\tau_{\textrm{real}}\in \mathcal{D}_{\textrm{eval}}} \frac{1}{T_{\textrm{min}}}\sum_{t = 1}^{T_{\textrm{min}}} | \delta_{c, t}^{\text{real}} - \delta_{c, t}^{\text{imag}} |$ where $\mathcal{D}_{\textrm{eval}}$ is an evaluation split, $T_{\textrm{min}} = \min(|\tau_{\textrm{real}}|, |\tau_{\textrm{imag}}|)$, and $\tau_{\textrm{imag}}$ is generated from $\tau_{\textrm{real}}$ .
For example, for a chasing entity, $\delta_{c, t}^{\text{real}}$ decreases as $t$ increases.  
If a model mistakenly predicts the entity to be immobile,  $\delta_{c, t}^{\text{imag}}$ remains a constant as $t$ progresses. 
In this case, $\Delta_{\text{dist}}$ is non-negligible, indicating an error.
All evaluation metrics are given in \autoref{tab:imagine}.  
 The ordering of the models is similar to that in the evaluation with ground-truth trajectories. 
 \ours is still superior to all baselines in all metrics.  

\subsection{Application: agents that discuss plans with humans}
\label{sec:application}

In this section, we showcase the practicality of our \lwm{} by illustrating that it can facilitate \textit{plan discussions} between an agent and a human supervisor.
This approach has the potential to improve the transparency, safety, and performance of real-world agents.
\looseness=-1

We imagine an agent ordered to perform a task in a previously unseen environment (\autoref{fig:application}). 
Letting the agent perform the task immediately would be extremely risky because of its imperfect knowledge of the environment. 
Implementing a world model enables the agent to imagine a solution trajectory and present it to a human as a \textit{plan} for review.
Conveying plans as trajectories helps the human envision the future behavior of the agent in the real world.
Furthermore, the human can improve this behavior by providing feedback to enhance the policy that produces the plan.

A human can update the policy by telling the agent which actions it should have taken.
This type of feedback can be incorporated using some form of imitation learning.
An agent equipped with a \lwm additionally enables the human to \textbf{update its policy by giving language feedback that aims to modify its world model}. 
Although an observational world model also allows this form of adaptation, it requires much more effort from the human to generate the feedback.
Concretely, the human has to generate observations in the same format as those in the agent's plan (e.g., they have to draw grids in this setting). 
Furthermore, many abstract concepts may not be efficiently or precisely specified through non-verbal communication. \looseness=-1

We simulate this scenario by placing agents with randomly initialized policies in test environments.
These agents are forbidden to interact with the environments. 
However, they are equipped with world models, which allows for imaginary policy update.
The world models are the ones we evaluated in the previous section.
Importantly, the models were not trained on any data collected in the environments, simulating the fact that these environments are completely new to the agents.
\looseness=-1

We train all policies with imitation learning, considering two types of feedback: in \textit{online imitation learning} \citep{ross2011reduction}, the expert suggests the best actions to take in the states present in the plan; in the \textit{filtered behavior cloning} setting, the expert simply overwrites the agent's plan with their own plan.
In the latter setting, the agent chooses the plans that achieve the highest returns according to their world models to imitate. 
We experiment with a near-optimal expert and a suboptimal expert.
We provide more details in \autoref{sec:imitation}.

The agents endowed with \lwm{}s can also process language feedback aiming to change their world models.
This feedback is simulated by the game manuals accompanying the environments.
It serves as the input $\vec \ell$ of the \lwm{}s.
We suppose that a human gives this feedback once to an agent, before adapting it via imitation learning.
\looseness=-1

We present the performance of the agents after adaptation in \autoref{tab:application}.
Learning with the \none world model amounts to the case where the human provides only imitation-learning feedback and cannot adapt the world model via language.
Meanwhile, learning with \ours represents the case where the human can use language feedback to improve the world model. 
In all evaluation settings, we observe significant improvements in the average return of policies that adopt our \ours. 
There are still considerable gaps compared to using the \oracle model, indicating that our model still has room for improvement. 

\begin{table}[!htb]
    \setlength{\tabcolsep}{2pt}
    \scriptsize
    \caption{Average returns ($\uparrow$) in real environments of policies trained with imaginary imitation learning using world models. Bold numbers indicate the best non-oracle means in the corresponding settings. An expanded table with all models and details on how the metric was computed are available in \autoref{sec:extended-result}. \looseness=-1}
    \label{tab:application}
    \centering
    \begin{tabular}{lfggg}
    \toprule
    \rowcolor{white}
    & & NewCombo & NewAttr & NewAll \\
    \rowcolor{white}
    Setting & World model & (easy) & (medium) & (hard) \\
    \midrule
    \rowcolor{white}
    & \none & 0.75 \textcolor{gray}{\ssmall{$\pm$ 0.16}} & -0.41 \textcolor{gray}{\ssmall{$\pm$ 0.21}} & -0.21 \textcolor{gray}{\ssmall{$\pm$ 0.21}} \\
    \rowcolor{white}
    & \ours (ours) & \textbf{1.01} \textcolor{gray}{\ssmall{$\pm$ 0.12}} & ~\textbf{0.96} \textcolor{gray}{\ssmall{$\pm$ 0.17}} & ~\textbf{0.62} \textcolor{gray}{\ssmall{$\pm$ 0.21}} \\
    \multirow{-3}{*}{\shortstack[l]{Online IL \\ \textit{\scriptsize\textcolor{darkgray}{(near-optimal)}}}}
    & \oracle & 1.04 \textcolor{gray}{\ssmall{$\pm$ 0.13}} & ~0.85 \textcolor{gray}{\ssmall{$\pm$ 0.20}} & ~0.91 \textcolor{gray}{\ssmall{$\pm$ 0.18}} \\
    \addlinespace
    \rowcolor{white}
    & \none & 0.77 \textcolor{gray}{\ssmall{$\pm$ 0.14}} & -0.42 \textcolor{gray}{\ssmall{$\pm$ 0.15}} & -0.30 \textcolor{gray}{\ssmall{$\pm$ 0.16}} \\
    \rowcolor{white}
    & \ours (ours) & \textbf{1.18} \textcolor{gray}{\ssmall{$\pm$ 0.10}} & ~\textbf{0.75} \textcolor{gray}{\ssmall{$\pm$ 0.20}} & ~\textbf{0.44} \textcolor{gray}{\ssmall{$\pm$ 0.18}} \\
    \multirow{-3}{*}{\shortstack[l]{Filtered BC \\ \textit{\scriptsize\textcolor{darkgray}{(near-optimal)}}}}& \oracle & 1.17 \textcolor{gray}{\ssmall{$\pm$ 0.11}} & 0.84 \textcolor{gray}{\ssmall{$\pm$ 0.19}} & ~0.80 \textcolor{gray}{\ssmall{$\pm$ 0.18}} \\
    \addlinespace
    \rowcolor{white}
    & \none & 0.71 \textcolor{gray}{\ssmall{$\pm$ 0.15}} & -0.35 \textcolor{gray}{\ssmall{$\pm$ 0.18}} & -0.33 \textcolor{gray}{\ssmall{$\pm$ 0.17}} \\
    \rowcolor{white}
    & \ours (ours) & \textbf{0.98} \textcolor{gray}{\ssmall{$\pm$ 0.13}} & ~\textbf{0.29} \textcolor{gray}{\ssmall{$\pm$ 0.25}} & ~\textbf{0.13} \textcolor{gray}{\ssmall{$\pm$ 0.19}} \\
    \multirow{-3}{*}{\shortstack[l]{Filtered BC \\ \textit{\scriptsize\textcolor{darkgray}{(suboptimal)}}}} & \oracle & 1.09 \textcolor{gray}{\ssmall{$\pm$ 0.13}} & ~0.50 \textcolor{gray}{\ssmall{$\pm$ 0.24}} & ~0.49 \textcolor{gray}{\ssmall{$\pm$ 0.18}} \\

    \bottomrule
    \end{tabular}    
\end{table}
\section{Related work}

\paragraph{World models.} World models have a rich history dating back to the 1980s \citep{werbos1987learning}. 
The base architecture has evolved from feed-forward neural networks \citep{werbos1987learning}, to recurrent neural networks \citep{schmidhuber1990making,schmidhuber1990line,schmidhuber1991possibility}, and most recently, Transformers \citep{robine2023transformerbased,micheli2022transformers}. 
In RL settings, world models are the key component of model-based approaches, which train policies in simulation to reduce the amount of interactions with real environments. 
Model-based RL has been successful in a variety of robotic tasks \citep{finn2017deep} and video games \citep{hafner2019dream,hafner2020mastering,hafner2023mastering}. 
However, the incorporation of language information into world models has been underexplored.
\citet{cowen2020emergent} propose language-conditioned world models but focus on emergent language rather than human language.
\citet{poudel2023langwm} incorporate features language into the representations of the model.
These approaches, however, do not use language to control a world model.

\paragraph{Language-based adaptation.} Language information has been incorporated into various aspects of learning. 
In instruction following \citep{bisk2016natural,misra2018mapping,anderson2018vision,nguyen2019help}, agents are given descriptions of the desired behaviors and learn to interpret them to perform tasks. 
Language-based learning \citep{nguyen2021interactive,scheurer2023training} employs language-based feedback to train models.
Another line of work uses language descriptions of environment dynamics to improve policy learning \citep{narasimhan2018grounding, branavan2012learning, hanjie2021grounding, wu2023read, nottingham2022learning, Zhong2020RTFM}.
Rather than using texts to directly improve a policy, our work leverages them to enhance a model of an environment.
Recently, several papers propose agents that can read text manuals to play games \citep{wu2023read,wu2023spring}.
Our work differs from these papers in that we aim to build models that capture exactly the transition function of an environment. 

\paragraph{Compositional generalization for language-guided world models.} \citet{lin2024learning} model a variety of text-augmented environments but do not demonstrate the generalizability of their approach in \messenger.
Recent work \citep{zhao2022toward,du2024learning,zhou2024robodreamer,zhang2024combo} has developed \lwm{}s with compositional generalizability.
While these papers operate on more visually realistic domains than ours, the language they study is simpler, focusing on concepts that correspond to straightforward mappings from input to output such as colors and objects. 
In contrast, the concepts in \messenger are more intricate, regarding interactions among multiple entities. 
\looseness=-1
\section{Conclusion}
We introduce \textit{Language-Guided World Models}, which can be adapted through natural language. 
We outline numerous advantages of these models over traditional observational world models.
Our model is still lacking in performance and the grid-world environments we experiment with severely underrepresent the real world.
Nevertheless, we hope that this work helps envision the potential of \lwm{}s in enhancing the controllability of artificial agents and inspires future efforts to address the compositional generalization challenge.  
\section*{Acknowledgements}
We thank Ameet Deshpande, Vishvak Murahari, and Howard Chen from the Princeton NLP group for valuable feedback, comments, and discussions. We thank Kurtland Chua for helpful feedback. This material is based upon work supported by the National Science Foundation under Grant Nos. 2107048 and 2239363. Any opinions, findings, and conclusions or recommendations expressed in this material are those of the author(s) and do not necessarily reflect the views of the National Science Foundation.

\bibliography{journal_full,bib}

\clearpage
\appendix

% \onecolumn

\section{GPTHard model} \label{appendix:chatgpt}
This approach leverages the language-understanding capabilities of ChatGPT. 
Through few-shot prompting, we instruct this model to determine the identity of the entity mentioned in each manual description.
In this approach, we generate only the set of values $\vec m^{\textrm{val}}$ as in \autoref{eqn:attn-val}.
Instead of learning soft attention, we directly route the values to the identity embeddings.
Concretely, the feature vector added to $\vec i^c_t$ in \autoref{eqn:decoder-input} is $\vec z^c_t = \vec m^{\textrm{val}}_{j_c}$ where $j_c$ is the index of the description that mentions entity $c$ according to ChatGPT.

We compose the following prompt for parsing descriptions. We use the ``May 3, 2023'' release of ChatGPT. 
We feed to the model one description at a time instead of a whole manual of three descriptions. 
We ask it to also extract the role and movement pattern, but use only the parsed identity in the \gpthard model. 
The ``ChatGPT identity-parsing'' column in \autoref{tab:dataset} shows the fraction of games in each split in which ChatGPT correctly identifies all three identities in a game.
Note that the \oracle model uses the ground-truth parses rather than these parses.

\lstset{basicstyle=\footnotesize\ttfamily,breaklines=true}
\begin{lstlisting}
You are playing a role-playing video game where you will need to read textual descriptions to figure out the attributes of a character.
                                                                                
This is a list of characters and their corresponding IDs:                       
airplane: 2                                                                     
mage: 3                                                                         
dog: 4                                                                          
bird: 5                                                                         
fish: 6                                                                         
scientist: 7                                                                                                                                            
thief: 8                                                                           
ship: 9                                                                         
ball: 10                                                                        
robot: 11                                                                       
queen: 12                                                                       
sword: 13                                                                       
                                                                                
This is a list of movement types and their corresponding IDs:                   
chasing: 0                                                                      
fleeing: 1                                                                      
stationary: 2                                                                   
                                                                                
This is a list of role types and their corresponding IDs:                       
dangerous enemy: 0                                                              
secret message: 1                                                               
essential objective: 2                                                          
                                                                                
Now, read a description and tell me which character is being mentioned and what are its movement type and role type. Your answer should follow this format:
```                                                                  
Answer: Character ID, movement type ID, role type ID                            
```

Here are a few examples:                                                        
                                                                                
Description: the plane that's flying near where you are is the critical objective.
Answer: 2, 0, 2                                                                 
                                                                                
Description: the escaping humanoid is an important goal.                        
Answer: 11, 1, 2                                                                
                                                                                
Description: the mage is inching near you is a lethal opponent.                 
Answer: 3, 0, 0                                                                 
                                                                                
Description: the classified document is the hound coming your way.              
Answer: 4, 0, 1                                                                 
                                                                                
Description: the important goal is the orb which is creeping close to you.         
Answer: 10, 0, 2                                                                
                                                                                
Now provide the answer for the following description. Follow the format of the previous answers:
                                                                                
Description: [PLACEHOLDER]                                   

\end{lstlisting}

% ---------------
% DATASET DETAILS
% ---------------
\section{Dataset}\label{sec:appendix-dataset-details}
\label{sec:dataset}

\begin{table*}[t!]
\setlength{\tabcolsep}{5pt}
\scriptsize
\centering
\begin{tabular}{llcccc}
    \toprule
    \multicolumn{2}{l}{Split} & Unique games & Unique descriptions & Trajectories & ChatGPT identity-parsing accuracy (\%) \\
    \midrule
    \multicolumn{2}{l}{Train} & 1,536 & 986 & 101,376 & 92 \\
    \midrule
    \addlinespace
    & NewCombo & 896 & 598 & 4,480 & 89 \\
    & NewAttr & 204 & 319 & 1,020 & 88 \\
    \multirow{-3}{*}{Dev} & NewAll & 856 & 1,028 & 4,280 & 86 \\
    \midrule
     \addlinespace
    & NewCombo & 896 & 587 & 4,480 & 90 \\
    & NewAttr & 204 & 306 & 1,020 & 93 \\
    \multirow{-3}{*}{Test} & NewAll & 856 & 1,016 & 4,280 & 88\\
    \bottomrule
\end{tabular}
\caption{\messenger data statistics. The last column shows the fraction of games in each split in which ChatGPT correctly identifies all three identities in a game.} 
\label{tab:dataset}
\end{table*}

Statistics of our dataset are provided in~\autoref{tab:dataset}. The maximum trajectory length is 32. We implement five rule-based behavior policies: \texttt{survive} (avoid the enemy and goal), \texttt{win the game}, \texttt{suicide} (go to the enemy), \texttt{obtain the message}, and \texttt{act randomly}. 
The \texttt{survive} policy acts randomly when the distances to the enemy and the goal are greater than or equal to 6. 
Otherwise, it takes the action that makes its distance to those entities at least 3. 
If that is impossible, it chooses the action that maximizes the minimum distance to one of the two entities.
The \texttt{win the game} policy is not optimal: it simply aims to obtain the message and then run to the goal, without having a strategy to avoid the enemy.
We run a breadth-first search to find the next best action to get to an entity.

For the training split, we generate 66 trajectories per game. 
The behavior policy for each trajectory is chosen uniformly randomly among the five rule-based policies.
For each evaluation split, we generate 5 trajectories per game, using every rule-based policy to generate trajectories. 

\section{Training details}
\label{sec:training}
\begin{table}[ht!]
\setlength{\tabcolsep}{5pt}
\centering
\begin{tabular}{ll}
    \toprule 
    Hyperparameter & Value \\
    \midrule
    Hidden size & 256 \\
    Number of encoder layers & 4 \\
    Number of decoder layers & 4 \\
    Number of decoder token blocks & 33 \\
    Dropout rate & 0.1 \\
    Batch size & 32 \\
    Number of training batches & 100K \\
    Evaluation every & 500 batches \\
    Optimizer & AdamW \\
    Learning rate & 1e-4 \\
    Max. gradient norm & 10 \\
    \bottomrule
\end{tabular}
\caption{Training hyperparameters.} 
\label{tab:hyperparams}
\end{table}

Our implementation of Transformer is largely based on the IRIS codebase \citep{micheli2022transformers}.\footnote{\url{https://github.com/eloialonso/iris}}
We implement cross-attention for the \standard baseline, and EMMA for our model. 

\paragraph{Initialization.} 
We find that the default PyTorch initialization scheme does not suffice for our model to generalize compositionally. 
We adopt the following initialization scheme from the IRIS codebase:

\begin{minted}[fontsize=\scriptsize]{python}
def init_weights(module):                                                       
    if isinstance(module, (nn.Linear, nn.Embedding)):                              
        module.weight.data.normal_(mean=0.0, std=0.02)                          
        if isinstance(module, nn.Linear) and module.bias is not None:              
            module.bias.data.zero_()                                            
    elif isinstance(module, nn.LayerNorm):                                      
        module.bias.data.zero_()                                                
        module.weight.data.fill_(1.0)    
\end{minted} 
which is evoked by calling \texttt{self.apply(init\_weights)} in the model's constructor. 
We initialize all models with this scheme, but only \ours and \oracle perform well consistently on various random seeds.

\paragraph{Compute resources.}  
Experiments were primarily run on a cluster of NVIDIA RTX2080 GPUs, and each experiment was run on a single device. To generate Table \ref{tab:real-eval}, we trained each world model for 24 GPU hours, 5 seeds each. To generate Table \ref{tab:application} and \ref{tab:extended-application}, we trained each of the 5 world models on each of the 90 games (3 difficulties for 30 game configurations) using the 3 different downstream policy training strategies, with each game being 12 GPU hours. 

\section{Imitation learning experiments}
\label{sec:imitation}
The learning policy follows the EMMA-based policy architecture of \cite{hanjie2021grounding}, which at each time step processes a stack of 3 most recent observations with a convolution-then-MLP encoder.
We train the policy with 2,000 batches using the same optimizer hyperparameters as those of the world models. 

For the online IL setting, we use the \texttt{win the game} rule-based policy (\autoref{sec:dataset}) as the expert. For the filtered BC setting, we train an EMMA policy to overfit the test environment. 
We then use a fully converged checkpoint of the policy as the near-optimal expert, and a not fully converged checkpoint as the suboptimal expert. 
The former is trained for 10,000 iterations and the latter is trained for 2,000 iterations.

The test environments are randomly chosen from the test splits. 
We select 10 environments per split.
We evaluate each policy for 48 episodes in the real environment. 
These episodes cover all 24 initial configurations of a stage-two \messenger game.

\section{Extended results}
\label{sec:extended-result}

\begin{figure*}[t]
    \includegraphics[width=\textwidth]{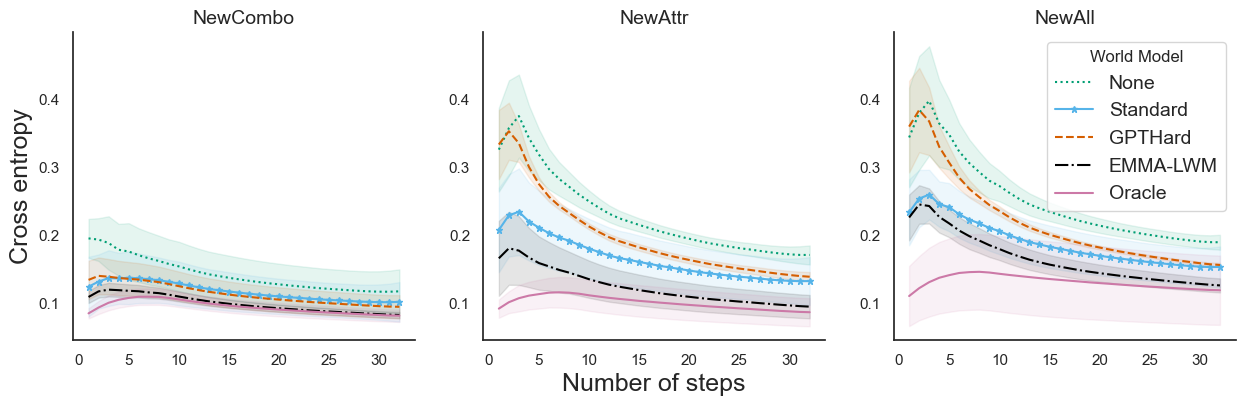}
    \caption{The cross entropy losses of the models when conditioned on ground-truth trajectory prefixes up to a certain length. We plot the means with 95\% t-value confidence intervals. The losses generally decrease as the prefix length increases. \ours outperforms baselines given any prefix length. \looseness=-1}
     \label{fig:in-context}
\end{figure*}

\autoref{fig:in-context} studies the performance of the models when conditioned on prefixes of the ground-truth trajectories.
The losses of all models decrease as the prefix length increases, but the baselines cannot close the gaps with \ours.
Across all splits, \ours conditioned on a one-step history outperforms \none conditioned on one third of a ground-truth trajectory, demonstrating that our model has effectively leveraged the textual information.

\autoref{tab:extended-application} presents the results of all the models in the simulation of plan discussion (\autoref{sec:application}).

\begin{table*}[t!]
    \footnotesize
    \caption{Average returns ($\uparrow$) in real environments of policies trained with imaginary imitation learning using world models. For each world model type, we use the best checkpoint of a run chosen randomly among the five runs mentioned in \autoref{tab:real-eval}. Experiments are conducted in 90 environments randomly chosen from the test splits (30 from each split). For each environment and learned policy, we compute the average return over 48 runs. For each split, we report the means of the average returns in the 30 environments with 95\% t-value confidence intervals. Bold numbers indicate the best non-oracle means in the corresponding settings. \ours outperforms all baselines in all settings. \looseness=-1}
    \label{tab:extended-application}
    \centering
    \begin{tabular}{lfggg}
    \toprule
    \rowcolor{white}
    & & NewCombo & NewAttr & NewAll \\
    \rowcolor{white}
    Setting & World model & (easy) & (medium) & (hard) \\
    \midrule
        \rowcolor{white}
    & \none & 0.75 \textcolor{gray}{\ssmall{$\pm$ 0.16}} & -0.41 \textcolor{gray}{\ssmall{$\pm$ 0.21}} & -0.21 \textcolor{gray}{\ssmall{$\pm$ 0.21}} \\
    \rowcolor{white}
    & \standard & 0.93 \textcolor{gray}{\ssmall{$\pm$ 0.13}} & ~0.04 \textcolor{gray}{\ssmall{$\pm$ 0.26}} & ~0.30 \textcolor{gray}{\ssmall{$\pm$ 0.22}} \\
    \rowcolor{white}
    & \gpthard & 0.82 \textcolor{gray}{\ssmall{$\pm$ 0.15}} & -0.20 \textcolor{gray}{\ssmall{$\pm$ 0.20}} & -0.06 \textcolor{gray}{\ssmall{$\pm$ 0.21}} \\
    \rowcolor{white}
    & \ours (ours) & \textbf{1.01 \textcolor{gray}{\ssmall{$\pm$ 0.12}}} & ~\textbf{0.96 \textcolor{gray}{\ssmall{$\pm$ 0.17}}} & ~\textbf{0.62 \textcolor{gray}{\ssmall{$\pm$ 0.21}}} \\
    \multirow{-5}{*}{\shortstack[l]{Online IL \\ \textit{\textcolor{darkgray}{(near-optimal expert)}}}}
    & \oracle & 1.04 \textcolor{gray}{\ssmall{$\pm$ 0.13}} & ~0.85 \textcolor{gray}{\ssmall{$\pm$ 0.20}} & ~0.91 \textcolor{gray}{\ssmall{$\pm$ 0.18}} \\
    \addlinespace
    \rowcolor{white}
    & \none & 0.77 \textcolor{gray}{\ssmall{$\pm$ 0.14}} & -0.42 \textcolor{gray}{\ssmall{$\pm$ 0.15}} & -0.30 \textcolor{gray}{\ssmall{$\pm$ 0.16}} \\
    \rowcolor{white}
    & \standard & 1.05 \textcolor{gray}{\ssmall{$\pm$ 0.14}} & ~0.20 \textcolor{gray}{\ssmall{$\pm$ 0.27}} & ~0.17 \textcolor{gray}{\ssmall{$\pm$ 0.20}} \\
    \rowcolor{white}
    & \gpthard & 0.79 \textcolor{gray}{\ssmall{$\pm$ 0.15}} & -0.10 \textcolor{gray}{\ssmall{$\pm$ 0.20}} & -0.07 \textcolor{gray}{\ssmall{$\pm$ 0.20}} \\
    \rowcolor{white}
    & \ours (ours) & \textbf{1.18 \textcolor{gray}{\ssmall{$\pm$ 0.10}}} & ~\textbf{0.75 \textcolor{gray}{\ssmall{$\pm$ 0.20}}} & ~\textbf{0.44 \textcolor{gray}{\ssmall{$\pm$ 0.18}}} \\
    \multirow{-5}{*}{\shortstack[l]{Filtered BC \\ \textit{\textcolor{darkgray}{(near-optimal expert)}}}}& \oracle & 1.17 \textcolor{gray}{\ssmall{$\pm$ 0.11}} & ~0.84 \textcolor{gray}{\ssmall{$\pm$ 0.19}} & ~0.80 \textcolor{gray}{\ssmall{$\pm$ 0.18}} \\
    \addlinespace
    \rowcolor{white}
    & \none & 0.71 \textcolor{gray}{\ssmall{$\pm$ 0.15}} & -0.35 \textcolor{gray}{\ssmall{$\pm$ 0.18}} & -0.33 \textcolor{gray}{\ssmall{$\pm$ 0.17}} \\
    \rowcolor{white}
    & \standard & 0.68 \textcolor{gray}{\ssmall{$\pm$ 0.15}} & -0.15 \textcolor{gray}{\ssmall{$\pm$ 0.21}} & -0.10 \textcolor{gray}{\ssmall{$\pm$ 0.17}} \\
    \rowcolor{white}
    & \gpthard & 0.75 \textcolor{gray}{\ssmall{$\pm$ 0.22}} & ~0.05 \textcolor{gray}{\ssmall{$\pm$ 0.25}} & ~0.06 \textcolor{gray}{\ssmall{$\pm$ 0.17}} \\
    \rowcolor{white}
    & \ours (ours) & \textbf{0.98 \textcolor{gray}{\ssmall{$\pm$ 0.13}}} & ~\textbf{0.29 \textcolor{gray}{\ssmall{$\pm$ 0.25}}} & ~\textbf{0.13 \textcolor{gray}{\ssmall{$\pm$ 0.19}}} \\
    \multirow{-5}{*}{\shortstack[l]{Filtered BC \\ \textit{\textcolor{darkgray}{(suboptimal expert)}}}} & \oracle & 1.09 \textcolor{gray}{\ssmall{$\pm$ 0.13}} & ~0.50 \textcolor{gray}{\ssmall{$\pm$ 0.24}} & ~0.49 \textcolor{gray}{\ssmall{$\pm$ 0.18}} \\
    \bottomrule
    \end{tabular}    
\end{table*}

\end{document}